\documentclass[10pt, a4paper]{article}

\usepackage[final]{lrec2026} % this is the new style
\title{SemiAdapt and SemiLoRA: Efficient Domain Adaptation for Transformer-based Low-Resource Language Translation with a Case Study on Irish}

\name{Josh McGiff$^{*}$ and Nikola S. Nikolov}

\address{
Department of Computer Science and Information Systems \\
University of Limerick, Ireland \\
\texttt{$^{*}$josh.mcgiff@ul.ie}
}

\abstract{
Fine-tuning is widely used to tailor large language models for specific tasks such as neural machine translation (NMT). However, leveraging transfer learning is computationally expensive when fine-tuning large multilingual models with billions of parameters, thus creating a barrier to entry for researchers working on low-resource domains such as Irish translation. Parameter-efficient fine-tuning (PEFT) bridges this gap by training on a fraction of the original model parameters, with the Low-Rank Adaptation (LoRA) approach introducing small, trainable adapter layers. We introduce SemiAdapt and SemiLoRA as semi-supervised inference-efficient approaches that strengthen domain adaptation and lead to improved overall performance in NMT. We demonstrate that SemiAdapt can outperform full-domain fine-tuning, while most notably, SemiLoRA can propel PEFT methods to match or even outperform full-model fine-tuning. We further evaluate domain-by-dataset fine-tuning and demonstrate that our embedding-based inference methods perform especially well on larger and noisier corpora. All Irish translation models developed in this work are released as open resources. These methods aim to make high-quality domain adaptation and fine-tuning more accessible to researchers working with low-resource languages. 
\newline  
}
\usepackage{makecell}
\usepackage{booktabs} 
\usepackage{hyperref}
\usepackage{xurl} % better line-breaking for URLs
\hypersetup{breaklinks=true} % allow links to wrap properly

\begin{document}
% Reduce paragraph spacing and indent first line
% \setlength{\parskip}{1pt}
% \setlength{\parindent}{1em}
\maketitleabstract

\section{Introduction}

Since their introduction, Transformer architectures \cite{vaswani2017attention} have reshaped the landscape of natural language processing (NLP), achieving substantial improvements in diverse language modeling tasks such as text generation \cite{lewis-etal-2020-bart}. Furthermore, Transformer-based architectures have revolutionised neural machine translation (NMT), particularly for low-resource languages (LRLs)\cite{nllb2024scaling}.

NMT is the process of using a neural network to encode a sentence in a source language and to decode a corresponding translated sentence in a target language \cite{bahdanau2014neural}. Although rule-based \cite{castilho2017neural,espana2016hybrid} and statistical machine translation \cite{bojar-etal-2015-findings, espana2016hybrid} has had historical success, NMT now dominates as the prevailing paradigm \cite{castilho2018approaches}. 

Despite constant advancements in the field of natural language generation, NMT remains the most extensively studied generative language modelling task in LRL communities \cite{mcgiff2025overcoming}. In an era where these advancements are empowering services such as ChatGPT and Claude, the prevalence of NMT in these language communities indicates that they are stuck playing catch-up. The domination of English language content on the internet\footnote{\href{https://w3techs.com/technologies/overview/content_language}{W3Techs: Usage of content languages for websites} (\url{https://w3techs.com/technologies/overview/content_language})}
 extends to the widely popular generative services available online \cite{robinson2023chatgpt}. This existing bias in the data enables research and development for majority languages such as English, and presents major challenges for less-represented languages \cite{mcgiff2025overcoming}. Given the large amount of unstructured data required to build coherent generative language models for a language, popular services such as ChatGPT are unable to accurately represent minority languages and their communities. A recent study found that popular large language models (LLMs) struggle with the syntactic complexities of LRLs such as Irish \cite{mcgiffirishblimp}. Furthermore, the general lack of literature evaluating LLMs for these languages indicates that these language communities are often an afterthought for LLM-powered tools and services.

In the context of Irish language translation, previous work has assessed various statistical and NMT methods on their capacity to translate from English to Irish and vice versa \cite{dowling2018smt, lankford2022human, defauw2019developing}. However, translating from English to Irish poses a greater challenge for large, multilingual models than the reverse direction \cite{tran2024irish,lankford2024gahealth}. This asymmetry arises because models such as NLLB-200 \cite{nllb2024scaling} are trained to generate English text across hundreds of language pairs, giving the decoder extensive exposure to English, while LRLs such as Irish are rarely used as target languages and are typically only decoded in the English to Irish direction \cite{costa2022no}. Consequently, the model’s decoder learns richer and more stable representations for English than for Irish, thus making generation into Irish more difficult. These issues are further compounded by the Irish language's morphological complexity, vocabulary sparsity, and limited parallel data \cite{10.1007/s10590-020-09253-x}. Addressing these challenges in English to Irish translation remains an under-explored research area.

As a result, we introduce SemiAdapt and SemiLoRA as two efficient approaches for improving English to Irish translation. SemiAdapt and SemiLoRA are semi-supervised methods that involve zero-shot domain assignment to training data, followed by either full-model fine-tuning or the training of low-rank adaptation (LoRA) adapters. At inference time, domain embedding centroids are used to efficiently assign domains. We find that SemiLoRA outperforms full-model fine-tuning on some domains and enables LoRA-based methods to achieve performance comparable to full-model fine-tuning. Although full-model fine-tuning outperforms SemiLoRA on some domains, SemiLoRA demonstrates that it can also benefit from our semi-supervised, inference-efficient strategy. 
Overall this paper produces:
\begin{itemize}
    \item \textbf{SemiAdapt} and \textbf{SemiLoRA} as two efficient, semi-supervised, and embedding-informed approaches that can outperform standard fine-tuning for English to Irish translation.
    \item A comparative assessment of \textbf{domain-based fine-tuning} using \textbf{SemiLoRA} and \textbf{LoRA}.
    \item A comparative assessment of \textbf{full-model domain fine-tuning} with and without \textbf{SemiAdapt}.
    \item A comparative assessment of translation performance with \textbf{dataset-level} versus \textbf{sentence-level} domain labelling.
    \item An analysis of the \textbf{impact of domain selection} on translation performance.
    \item A suite of \textbf{open-source English to Irish translation models}\footnote{\href{https://huggingface.co/collections/lrec2026-anonsubmission/semiadapt-and-semilora-efficient-domain-adaptation-nmt-68f23fd42c42ccf0c59fbdad}{Models available on Hugging Face (Anonymous account):} \url{https://huggingface.co/collections/lrec2026-anonsubmission/semiadapt-and-semilora-efficient-domain-adaptation-nmt-68f23fd42c42ccf0c59fbdad}}, including \textbf{SemiAdapt}, \textbf{SemiLoRA}, and \textbf{fully fine-tuned NLLB-200} baselines\footnote{\href{https://anonymous.4open.science/r/SemiAdapt-SemiLoRA-C630/README.md}{Code available on GitHub (Anonymous account):} \url{https://anonymous.4open.science/r/SemiAdapt-SemiLoRA-C630/README.md}}.
\end{itemize}

This paper aims to arm LRL modelling researchers with an evaluation of two novel and efficient fine-tuning approaches for English to Irish translation that could be extended to other tasks that benefit from domain adaptation. In contrast to assuming that dataset-level domain labels are sufficiently granular, our sentence-level domain label assignment results in better translation performance across domains. Ultimately, SemiAdapt and SemiLora aims to empower researchers working on LRLs to build language models such as NMT systems that are both efficient and robust across domains, thus reducing reliance on large labelled datasets and full-parameter fine-tuning.

\section{Related Work}
\subsection{Multilingual Models}
LLMs and their capacity to generalise across multiple languages at once have been exploited to enhance language generation performance across tasks such as LRL translation \cite{mcgiff2025overcoming,  cahyawijaya_indonlg_2021, wongso_many--many_2023, tanwar_translating_2020, liu_low-resource_2022}. Furthermore, building language models with groups of linguistically-related languages has been equated with augmenting the training dataset by 33\% \cite{cahyawijaya_indonlg_2021}. Although some research suggests that smaller multilingual models can be outperformed by monolingual models for LRLs such as Slovene, multilingual models tend to perform better at scale \cite{noauthor_frontiers_nodate}. However, further research is required to both verify this generalisation for different language groups and to identify the inflection point where multilingual translation models outperform translation models for a given pair of languages.

\subsection{Neural Machine Translation}
In terms of Irish language machine translation research, Irish is mostly paired with English for training models \cite{dowling2018smt, lankford2022human, defauw2019developing}. However, given that training language models of linguistically-related languages can boost modelling performance \cite{cahyawijaya_indonlg_2021}, Irish could be better aligned with other Goidelic languages such as Scottish Gaelic and Manx \cite{anderson2024goidelex}. To our knowledge, there are no published research efforts exploring large Goidelic language family-based models. That said, code-mixing of English and Irish is very common in Ireland, with both languages often being used in the same sentence \cite{laoire2016irish}. The pervasive influence of English on Irish, a legacy of Ireland’s colonial and postcolonial linguistic history, has led to frequent code-mixing and borrowing in modern Irish language use \cite{hickey2009code}. Consequently, the strong relationship between the languages necessitates the integration of English in Irish translation systems, and enables English to Irish NLP in general.

Despite the fact that Irish is regarded as a LRL \cite{mcgiff2025overcoming,nllb2024scaling}, a growing body of work has explored machine translation for the language, spanning statistical, recurrent, and transformer-based approaches. During the mid-2010s, some work focused on enhancing statistical machine translation (SMT) methods for English to Irish translation \cite{arcan2016iris}. The first NMT systems for the language pair were produced in 2018, with their LSTM-based results underperforming compared to SMT models at the time \cite{dowling2018smt}. 

More recent work \cite{dowling2020human, defauw2019developing, lankford2024gahealth, lankford2024transformers} has been dominated by the advent of the Transformer architecture \cite{vaswani2017attention}. Two approaches have since used the OpenNMT ecosystem with the Transformer architecture as a base \cite{dowling2020human, defauw2019developing}.  Defauw \textit{et al.} explore the positive impact of back-translation of domain-specific monolingual data on translation performance. Additionally, they also indicate that misalignment detection-based filtering of synthetic sentence pairs can produce higher BLEU scores \cite{defauw2019developing}. Although Defauw \textit{et al.} explore domain-specific translation, their approach only focuses on the legal domain and infers domain labels at a corpus level. Other papers focus on specific domains such as health \cite{lankford2024gahealth} and legislation without exploring parameter-efficient fine-tuning (PEFT) such as LoRA \cite{lankford2024transformers}, or a wider set of domains at once. 

In terms of English to Irish translation performance, it is challenging to compare existing studies as many of them report results such as BLEU on different datasets and domains. Dowling \textit{et al.} report BLEU scores between 31.9-33.9 across various NMT setups on 1,500 random in-domain sentences sourced from datasets mostly found on the OPUS platform \cite{dowling2020human}. Langford \textit{et al.} achieve BLEU scores between 53.4 and 60.5 when exploring the effect of byte-pair encoding vocabulary size on Transformer-powered translation \cite{lankford2022human}. Tung \textit{et al.} highlight that English to Irish translation is poor on Llama 2-13B as a baseline with a BLEU score of 3.25 \cite{tran2024irish}. However, their approach (UCCIX) of fine-tuning LLama 2-13B for the pair achieves a BLEU score of 33.34 on the 500 LoResMT English to Irish evaluation set. 

Alternatively, Dowling \textit{et al.} perform human evaluation on OpenNMT-based English to Irish translations \cite{dowling2020human}. Surprisingly, their study was the first to include professional translators in evaluating English to Irish machine translated text. They found that a small group of four translators indicated that NMT was the most accurate in comparison with SMT methods, even when this perception did not fully align with their post-editing experience or fluency preferences. Their findings support the idea in generative language modelling that automatic scores do not necessarily reflect human experience or preference \cite{mathur2020tangled, escribe2019human}.

\subsection{Low-Resource Challenges}
While the term “low-resource” typically refers to a lack of data for training language models, it also extends to limitations in computational capacity and access to skilled researchers \cite{ogueji2021small}. A recent systematic review on language modelling for LRLs found that pre-training from scratch was rare, with the majority of studies opting to fine-tune models or develop prompt-engineering strategies for existing models such as ChatGPT \cite{mcgiff2025overcoming}. This indicates that limited access to computational resources could pose a barrier to entry for researchers wishing to focus on minority languages for their communities. 

Given these computational barriers, LoRA offers a parameter-efficient approach that can significantly reduce the computational requirements for fine-tuning language models \cite{hu2022lora}. LoRA works by freezing pre-trained model weights and injecting trainable low-rank decomposition matrices into selected target modules such as the self-attention layers \cite{zhang2023machine}. The original LoRA approach reduces the GPU memory required to fine-tune GPT-3 175B to roughly one-third of the original demand, while decreasing the number of trainable parameters by a factor of 10,000 \cite{hu2022lora}. LoRA does not necessarily compromise on performance either with the model quality being similar or better when fine-tuning models such as RoBERTa and GPT-3. As a result, LoRA offers the often computationally-constrained LRL researchers an approach to fine-tuning larger parameter models or many models in parallel on the same hardware.   

Furthermore, the modular design of LoRA adapters allows them to be swapped in and out without altering the underlying base model \cite{hu2022lora}. The modular design pairs well with domain adaptation as some form of domain detection can be used to route inputs to an appropriate LoRA adapter, thus enabling efficient adapter swapping without the need to load multiple large-parameter models during inference. Although, some studies introduce different complexities and inefficiencies by training models to predict the correct adapter or domain for a given input \cite{tian2025adapters,feng2024mixture}. In terms of NMT, numerous studies have empirically shown that domain adaptation improved translation performance \cite{freitag2016fast, chu2017empirical}, particularly for LRLs \cite{marashian-etal-2025-priest}. As a result, LoRA adapters could be used to bridge the gap in computational constraints for researchers looking to build NMT models for LRLs. 

\section{Experimental Setup}
\subsection{Data}
We source parallel sentences for English to Irish translation from a variety of sources, using the OPUS platform \cite{TIEDEMANN12.463}. Furthermore, we mine 813k additional sentence pairs from alternative sources such as the State Examinations Commission and Foclóir \cite{sec_exam_papers, focloir_ie}. We automatically extracted parallel English to Irish text from Irish school exam papers by aligning corresponding English and Irish language PDF versions at the page and text-block level using PyMuPDF\footnote{\href{https://github.com/pymupdf/PyMuPDF}{PyMuPDF:} \url{https://github.com/pymupdf/PyMuPDF}}. Text blocks were normalised by position and matched based on spatial proximity. We excluded language subjects as bilingual copies do not exist or contain irrelevant languages as opposed to English or Irish text content. This new dataset almost triples the existing OPUS-sourced sentence count and brings the total dataset to 1.32 million sentences. We report additional dataset-based features in Table~\ref{tab:datasets} 

Several studies reveal that web-crawled corpora contain useless text for LRLs, with only 59\% of Irish text in mC4 being deemed correct and natural language \cite{kreutzer2022quality, caswell2020language}. Consequently, we exclude web-crawled English to Irish text data from our experiments in order to maintain the linguistic integrity and naturalness of the material.   

It can be noted that there are on average 11.92 tokens per English sentence and 12.89 tokens per Irish sentence in the dataset. Interestingly, Irish sentences are about 8\% longer than corresponding English sentences.  

The general domain dominates as the largest domain with 813k sentence pairs sourced from an official Irish-English bilingual dictionary and Irish exam papers \cite{focloir_ie,sec_exam_papers}. We find that the vast majority of corpora are either related to Irish or European-level legislation, and  therefore are considered in the legal domain. DGT, EUConst, Gaois, DECP and EuBookshop are the five sources of legal text content, thus making the legal domain the second biggest domain with 478k sentence pairs. Encyclopedic and medical domains follow suit with 19k and 8.6k sentences respectively.

\begin{table}[!ht]
\begin{center}
\scriptsize
\setlength{\tabcolsep}{2pt}
\begin{tabularx}{\columnwidth}{|l|l|l|l|X|}
\hline
\textbf{Dataset} & \textbf{Sent.} & \textbf{EN} & \textbf{GA} & \textbf{Domain} \\
\hline
\makecell[l]{DGT \\ \tiny{\cite{TIEDEMANN12.463}}} & 173k & 4.1M & 4.4M & Legal \\
\hline
\makecell[l]{EUConst \\ \tiny{\cite{TIEDEMANN12.463}}} & 6.7k & 0.14M & 0.14M & Legal \\
\hline
\makecell[l]{Gaois \\ \tiny{\cite{gaois_legislation_2021}}} & 99k & 1.5M & 1.6M & Legal \\
\hline
\makecell[l]{Wikimedia \\ \tiny{\cite{TIEDEMANN12.463}}} & 19k & 0.49M & 0.52M & Encycl. \\
\hline
\makecell[l]{LoResMT \\ \tiny{\cite{ojha-etal-2021-findings}}} & 8.6k & 0.13M & 0.15M & Medical/COVID-19 \\
\hline
\makecell[l]{DECP \\ \tiny{\cite{koehn-2005-europarl}}} & 103k & 1.0M & 1.1M & Legal \\
\hline
\makecell[l]{EUBookshop \\ \tiny{\cite{skadicnvs2014billions}}} & 96k & 2.2M & 2.3M & Legal \\
\hline
\makecell[l]{SEC \\ \tiny{\cite{sec_exam_papers}}} & 212k & 2.5M & 2.7M & General \\
\hline
\makecell[l]{Foclóir \\ \tiny{\cite{focloir_ie}}} & 601k & 3.7M & 4.1M & General \\
\hline
\textbf{Total} & \textbf{1.32M} & \textbf{15.7M} & \textbf{17.0M} & --- \\
\hline
\end{tabularx}
\caption{English to Irish parallel corpus statistics. Sentence counts shown in thousands (k) with English (\textbf{EN}) and Irish (\textbf{GA}) token counts displayed in millions (M). }
\label{tab:datasets}
\end{center}
\end{table}

\subsection{Base Model}
All our experiments are conducted on nllb-200-distilled-600M \cite{nllb2024scaling} as a base model. This model is the smallest of the public NLLB-200 checkpoints released by Meta. We chose this base model to explore how a relatively small multilingual LLM can perform under a do-more-with-less approach. This configuration enables experimentation on modest hardware, thus offering practical insights for LRL researchers and communities operating under resource constraints. The models were fine-tuned using the Hugging Face transformers framework \cite{wolf-etal-2020-transformers} across two NVIDIA A100 80GB GPUs.

\section{Experiments}
\subsection{SemiLoRA: Semi-Supervised LoRA Fine-Tuning}
This experiment involves collating the training text content from the sources mentioned in Table~\ref{tab:datasets} and using a zero-shot natural language inference classifier to assign domain labels on a sentence level. 90\% of the available English to Irish parallel data are used for fine-tuning and the remaining 10\% are reserved for evaluation. Meta's bart-large-mnli model \cite{lewis-etal-2020-bart} enables noise reduction when creating domain-based dataset splits as not every sentence pair is particularly suited to the general domain of its source. We hypothesise that this preprocessing technique will form better domain groupings and consequently improve domain adaptation performance. This hypothesis is explored further in Section~\ref{sec:domainbydataset}.

% put img here of experiment flow

We define the domains for this experiment as general, legal, medical/COVID-19 and wiki/news. We experiment with two different approaches for domain classification with the zero-shot classifier. We create the first training split by providing the zero-shot classifier with all four domains. Alternatively, we create a second split using the same classifier but we exclude the general domain and apply a confidence threshold of 0.45 to the other domains; if no domain exceeds this threshold, the instance is assigned to the general domain.

The English sentences are assigned a domain using the zero-shot classifier  and the parallel dataset is subsequently split by domain. The data undergo a process of deduplication, randomisation, and line splitting to account for rows containing multiple sentences. Finally, the training dataset splits are tokenised using the nllb-200-distilled-600M tokeniser \cite{nllb2024scaling}. 

% Training stuff for NLLB & LoRA - General used as a base

We initially select the general domain as the starting point for training as NLLB-200 performs poorly on English to Irish translation by default. That is, we fine-tune entirely on the general domain split for each configuration. General domain fine-tuning was performed using the default \texttt{AdamW} optimiser provided in the \texttt{Transformers} library \cite{wolf-etal-2020-transformers} with a learning rate of \(5\times10^{-5}\). Training ran for three epochs with a per-device batch size of~4 and gradient accumulation over~4~steps.

We subsequently fine-tune LoRA adapters for each domain, with all the adapters for a single configuration fitting and simultaneously training on a single A100 GPU. For the LoRA-based fine-tuning experiments, we applied low-rank adapters where the configuration used a rank of \(r = 16\), scaling factor \(\alpha = 16\), and a dropout rate of~0.1. Adapters were inserted into the attention projection layers (\texttt{q\_proj}, \texttt{k\_proj}, \texttt{v\_proj}, \texttt{out\_proj}) as well as the feed-forward layers (\texttt{fc1}, \texttt{fc2}). Bias terms were kept frozen, and the task type was set to \texttt{SEQ\_2\_SEQ\_LM} to support encoder-decoder training. Fine-tuning was performed using the \texttt{AdamW} optimiser with a learning rate of \(1\times10^{-4}\), weight decay of~0.01, and mixed-precision (FP16) training for efficiency. Each model was trained for three epochs with a per-device batch size of~4 and gradient accumulation over~4~steps.

\begin{figure}[htbp]
        \centering
        \includegraphics[width=0.5\textwidth]{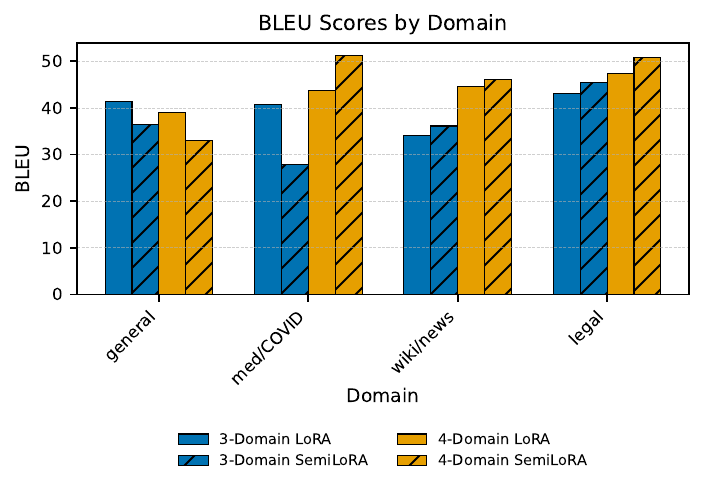}
        \caption{Comparison of BLEU scores by domain for LoRA and SemiLoRA models trained on three and four domains. }
        \label{fig:semilora}
    \end{figure}

%  need to write about the Semi part - how does it compare
The evaluation step is where LoRA and SemiLoRA diverge. SemiLoRA involves computing centroid embeddings for each domain in the training split and comparing them with the embeddings of input sentences in the evaluation dataset. Unlike the LoRA approach of reusing the zero-shot classifier to label input sentences, SemiLoRA is semi-supervised and only uses a classifer to label training data. SemiLoRA does not enlist a classification model at inference time. Specifically, the input embeddings are compared to each domain centroid using cosine similarity and assigned a label depending on the closest domain. Domains are essentially assigned based on the semantic similarity of a given input sentence and the average representation for each domain label. We report the results of these LoRA-based experiments in Table~\ref{tab:bleu-results} and Figure~\ref{fig:semilora}.

SemiLoRA with domains assigned by the zero-shot classifier outperforms the other approaches on each of the specific domains. Most notably, SemiLoRA achieves substantial improvements over LoRA, with gains of 11 BLEU in the medical domain, 1.7 BLEU in the wiki/news domain, and 3.5 BLEU in the legal domain. Results vary by method for defining domains. The four domains classified without a confidence threshold results in the best results for domain-specific translation. 

On the other hand, the three-domain approach where the general domain is labelled based on a confidence threshold, outperforms the domain-specific models on the general domain. This was somewhat surprising given that the confidence threshold-informed general domain split was half the size of the general domain chosen by the zero-shot classifier. This likely indicates that the SemiLoRA method of assigning domains by comparing input embeddings to domain embedding centroids, struggles with noisy general data. Combining a base model fine-tuned on the entire dataset with SemiLoRA for domain adaptation requires further exploration. Although, it is clear that the SemiLoRA semantic similarity approach can outperform the supervised approach of using a classifier at inference time, if suitable domain groupings are selected.

\subsection{SemiAdapt: Semi-Supervised Full-Model Fine-Tuning}

This experiment follows a similar setup to the previous SemiLoRA experiment, where full-model fine-tuning, as opposed to adapter fine-tuning, is completed on each domain. Similarly to SemiLoRA, SemiAdapt follows a semi-supervised approach by using domain labels sourced from Meta's zero-shot classification model \cite{lewis-etal-2020-bart} for training and by subsequently using domain embedding centroids to efficiently determine domain labels at inference time. This experiment also explores the impact of the number of domains on SemiAdapt performance. However, unlike SemiLoRA, the alternative SemiAdapt approach requires training all of the model’s learnable parameters and therefore cannot support multiple models being trained on a single A100 GPU.

Similarly to SemiLoRA, we choose the general domain as a base for fine-tuning. In other words, we fine-tuned on the general domain before fine-tuning on specific domains such as legal and medical domains. SemiAdapt fine-tuning was performed using the default \texttt{AdamW} optimiser provided in the \texttt{Transformers} library with a learning rate of \(5\times10^{-5}\). Training ran for three epochs with a per-device batch size of~4 and gradient accumulation over~4~steps. Unlike the SemiLoRA configuration, which used PEFT to adapt only low-rank adapter layers, the models in this experiment were fully fine-tuned. This means all model parameters were updated during training.

\begin{figure}[htbp]
        \centering
        \includegraphics[width=0.5\textwidth]{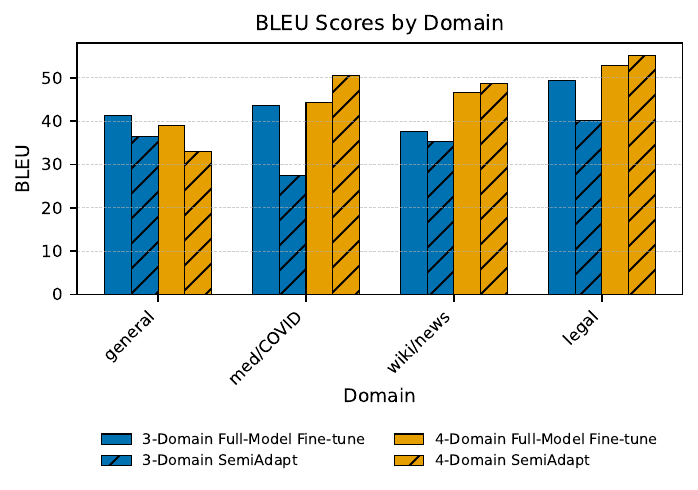}
        \caption{BLEU scores by domain comparing full-model fine-tuning and SemiAdapt models trained on three and four domains.}
        \label{fig:semiadapt}
    \end{figure}

\begin{figure}[htbp]
        \centering
        \includegraphics[width=0.5\textwidth]{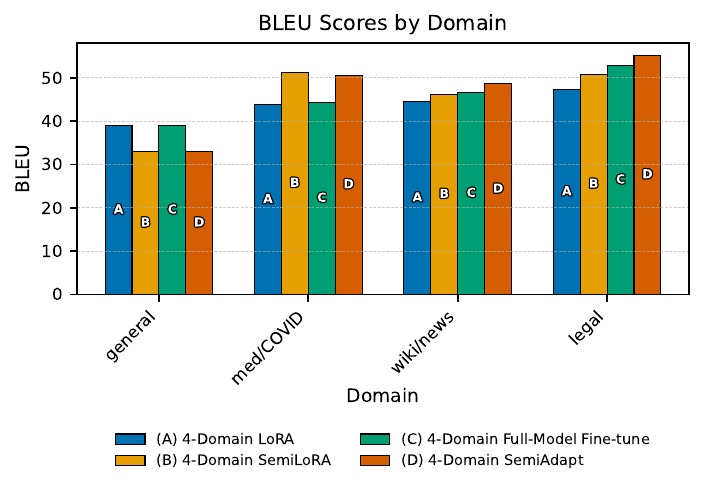}
        \caption{BLEU scores by domain comparing full-model fine-tuning and LoRA-based models trained on four domains.}
        \label{fig:compare4domain}
    \end{figure}

The evaluation step is where full-model fine-tuning and SemiAdapt diverge. Full-model fine-tuning is evaluated on an evaluation dataset that has been labelled with a zero-shot classification model, whereas SemiAdapt uses domain embedding centroids to apply labels at inference time. We include this experiment to explore if regular full-model fine-tuning can also benefit from semi-supervised domain assignment, as seen with LoRA-based models. The inclusion of this full-model fine-tuning setup helps to measure the upper bound of domain-specific translation performance and to provide a comparison point against PEFT methods such as LoRA.  We report the results of these full-model fine-tuning experiments in Table~\ref{tab:bleu-results} and Figures~\ref{fig:semiadapt} and ~\ref{fig:compare4domain}.

The results show that four-domain SemiAdapt outperforms the other approaches on each specific domain. Most notably, SemiAdapt achieves considerable improvements over full-model fine-tuning with zero-shot domain  classification at inference time on the same medical domain split with a 6.3 increase in BLEU score. More moderate improvements are achieved on wiki/news and legal domains with increases of 2 and 2.4 BLEU respectively. Although, it performs the worst on the general domain, thus indicating that this semi-supervised domain adaptation approach, in addition to SemiLoRA, could benefit from a base model fine-tuned on the entire dataset. 

Overall, these results reveal that SemiAdapt-based semi-supervised domain adaptation can also improve translation performance when performing full-model fine-tuning with NLLB-200. However, the substantially higher computational cost of full-model fine-tuning, compared to LoRA-based approaches such as SemiLoRA, may limit its practicality for researchers working with constrained resources or minority language settings. These results justify experimenting with SemiAdapt on language models for other downstream tasks such as question answering.  

Additionally, inter-experiment analysis reveals that SemiAdapt consistently surpasses SemiLoRA and LoRA in the wiki/news and legal domains. However, SemiAdapt falls slightly short (0.7 BLEU) of SemiLoRA on the medical domain, despite the SemiLoRA adapter being 1.39\% (8.65M parameters) of the NLLB-200 model's full size (623.72M parameters). Therefore, it is worth noting that the semi-supervised approach helps bridge the gap between LoRA and full-model fine-tuning, with SemiLoRA almost on par with fine-tuning entirely on the wiki/news domain, and less than two BLEU short of the legal domain. Not only does this indicate that SemiLoRA can help PEFT compete with full-model fine-tuning, but it also highlights that a semi-supervised embedding-based approach can cut out the need for a traditional classifier at inference time. 

As a result, SemiLoRA presents a particularly suitable approach for resource-constrained domains such as LRL modelling. Alternatively, SemiAdapt presents an inference-efficient approach that can improve domain adaptation and general translation performance for regular fine-tuning. Additionally, this approach could be applied to identify domain groupings within out-of-domain datasets, thereby extending the benefits of domain adaptation reported in the SemiLoRA and SemiAdapt experiments.

\subsection{Domain-by-Dataset Fine-Tuning} \label{sec:domainbydataset}
In this experiment, we explore our hypothesis from Experiment 4.1 where we suggest that sentence-level domain labelling could form better domain groupings and potentially improve domain adaptation performance. The previous SemiLoRA and SemiAdapt experiments involve splitting the dataset by domain using a zero-shot classifier on each sentence. 

In this experiment, we infer and assign domain labels on a corpus level. In other words, we assume every sentence pair in a corpus reflects the domain label identified in Table~\ref{tab:datasets}. 

For example, the LoResMT English to Irish dataset related to COVID-19 \cite{ojha-etal-2021-findings} contains sentences such as:

\begin{quote}
EN: \textit{What is added by this report?}

GA: \textit{Cad a chuireann an tuarascáil seo leis?}
\end{quote}

Although this example sentence is sourced from a medical/COVID-19 dataset, it does not specifically or exclusively apply to the medical domain. The sentence is more suited to a general, domain-agnostic label. Therefore, these inaccuracies support our hypothesis that sentence-level labeling could reduce noise in the dataset splits and consequently improve domain adaptation and translation performance.

The exam and dictionary corpora are assigned the general domain due to the out-of-domain nature on their text content. We merge all legal sources into a single legal corpus and assume that the Wikimedia and LoResMT datasets correspond to the wiki/news and medical domains, respectively. This time we use 90\% of the available English to Irish parallel data from each domain split for fine-tuning and reserve the remaining 10\% for evaluation. 

\begin{figure}[htbp]
        \centering
        \includegraphics[width=0.5\textwidth]{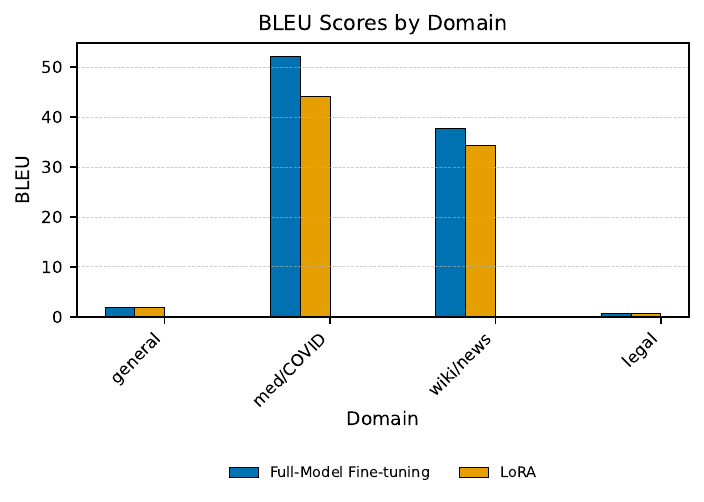}
        \caption{BLEU scores by domain comparing full-model fine-tuning and  LoRA-based models on the domain-by-dataset split.}
        \label{fig:DOMAINBYDATASET}
    \end{figure}

Unlike the SemiLoRA and SemiAdapt approaches, the domain is inferred by the dataset source and therefore does not require any classification process. We experiment with full-model fine-tuning and with using LoRA to do PEFT on each domain. We subsequently reuse the training parameters for the full-model fine-tuning and LoRA-based approaches from the previous experiments. The results of these domain-by-dataset experiments are reported in Table~\ref{tab:bleu-results} and Figure~\ref{fig:DOMAINBYDATASET}.

% Talk about the results
The results of this experiment indicate that model performance varied widely across domains when domain labels were assigned at a dataset level. We report that there is a negative correlation between dataset size and BLEU score performance. Both full-model fine-tuning and LoRA-based training methods perform best on the medical and wiki/news domains, i.e. on the domains with the least amount of sentence pairs. Although the full-model fine-tuning approach achieves BLEU scores of 52 and 38 on the medical and wiki/news domains respectively, its poor performance on the significantly larger general and legal domains suggests the model memorises domain-specific phrases as opposed to learning generalisable translation mappings within the domain itself. This suggests that large, noisy datasets could benefit from a more granular approach to domain labelling and grouping. This suggestion is supported by our results from the SemiLoRA and SemiAdapt experiments, where better translation performance was recorded across each of the domains.

\subsection{Full-Dataset Fine-Tuning (Baseline)}
\begin{figure}[htbp]
        \centering
        \includegraphics[width=0.5\textwidth]{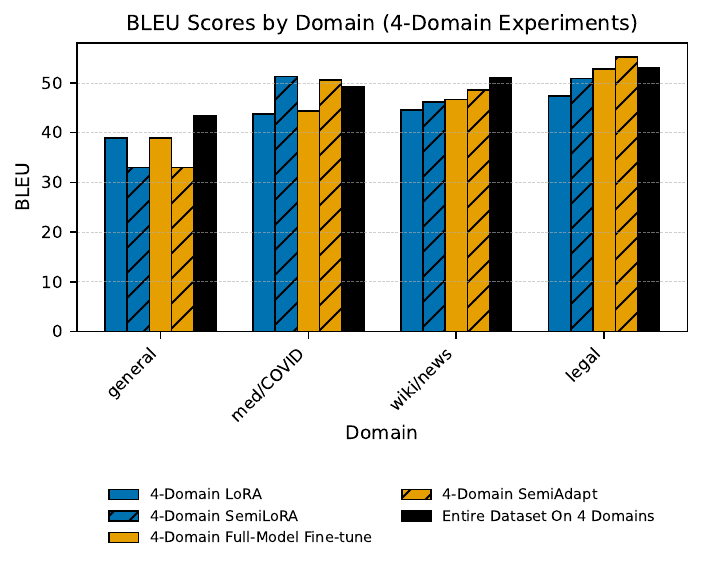}
        \caption{BLEU scores by domain comparing full-dataset fine-tuning, full-model domain fine-tuning, and LoRA-based models on the four-domain split.}
        \label{fig:BLEU_Comparison_4Domain_ColorBlindSafe}
    \end{figure}

This experiment involves combining each domain split from the four-domain SemiLoRA and SemiAdapt experiment training dataset. Unlike our previous experiments, the NLLB-200 base model is fine-tuned directly on this combined, domain-agnostic collection of English to Irish parallel sentences. We also reuse the evaluation set from the previous four-domain experiments, as this enables a direct comparison between this full-dataset fine-tuning approach and previous experiment methods. We use the same model parameter configuration as the aforementioned SemiAdapt experiment. We report the results of this full-dataset fine-tuning experiment and a comparison of other methods in Table~\ref{tab:bleu-results} and Figure~\ref{fig:BLEU_Comparison_4Domain_ColorBlindSafe}.

\begin{table}[h!]
\centering
\scriptsize
\begin{tabular}{lcccc}
\toprule
Model & General & Medical & WikiNews & Legal \\
\midrule
\multicolumn{5}{l}{\textbf{3-domain models}} \\
LoRA & 41.35 & 40.76 & 34.18 & 43.20 \\
Full-model domain FT & 41.35 & 43.67 & 37.54 & 49.30 \\
SemiLoRA & 36.49 & 27.78 & 36.13 & 45.36 \\
SemiAdapt & 36.49 & 27.50 & 35.20 & 40.16 \\
\addlinespace
\multicolumn{5}{l}{\textbf{4-domain models}} \\
LoRA & 38.94 & 43.77 & 44.54 & 47.35 \\
Full-model domain FT & 38.94 & 44.37 & 46.64 & 52.81 \\
SemiLoRA & 32.96 & 51.29\textbf{\textsuperscript{\dag}}
 & 46.18\textbf{\textsuperscript{\dag}} & 50.85\textbf{\textsuperscript{\dag}} \\
SemiAdapt & 32.96 & 50.61 & 48.59 & \textbf{55.21} \\
\addlinespace
\multicolumn{5}{l}{\textbf{Domain-by-dataset}} \\
LoRA & 1.87 & 44.20 & 34.28 & 0.58 \\
Full-model fine-tune & 1.87 & \textbf{52.18} & 37.79 & 0.62 \\
\addlinespace
\multicolumn{5}{l}{\textbf{Full-dataset fine-tuning}} \\
Full-model fine-tune & \textbf{43.33} & 49.15 & \textbf{51.08} & 53.15 \\
\addlinespace
\bottomrule
\end{tabular}
\caption{BLEU scores by domain for different fine-tuning configurations. Bold indicates the best score per domain. \textbf{\textsuperscript{\dag}} marks parameter-efficient models performing within 5 BLEU points of the best model for that domain.}
\label{tab:bleu-results}
\end{table}

The results show that full-dataset fine-tuning yields the strongest performance on the general and wiki/news domains. In contrast, both SemiLoRA and SemiAdapt outperform it on the medical domain, with SemiAdapt also leading on the legal domain. Nevertheless, LoRA and full-domain fine-tuning remain within five BLEU points of the full-dataset fine-tuning baseline on the general domain, and SemiAdapt is similarly close on wiki/news. 

These results indicate that full-dataset fine-tuning can be outperformed by domain-specific fine-tuning. Moreover, they show that SemiLoRA can either surpass or closely match the performance of full-model fine-tuning on both the entire dataset and domain-specific splits. This finding suggests that SemiLoRA offers a practical advantage for LRL research, as it enhances PEFT methods to achieve performance comparable to, or even exceeding, that of full-model fine-tuning. However, the strong performance of the full-dataset fine-tuning approach on the general domain supports our earlier argument that SemiLoRA and SemiAdapt could benefit from using a full-dataset fine-tuned model as a foundation for further adaptation. 

\section{Conclusion}
In this paper, we introduced two novel domain adaptation techniques, SemiLoRA and SemiAdapt, in addition to a suite of open-source models for English to Irish translation. We demonstrated that semi-supervised sentence-level domain adaptation enables LoRA-based models to match the performance of full-model fine-tuning. This in itself should empower resource-constrained LRL researchers to train language models that compete with full-sized models and to train multiple adapters at once in parallel. We have also shown that SemiAdapt outperforms standard domain-based fine-tuning, suggesting that leveraging domain embedding centroids for sentence-level domain labeling can enhance full-model fine-tuning as well as parameter-efficient approaches such as SemiLoRA.

For future work, we plan to extend our investigation of SemiLoRA and SemiAdapt to other LRLs and additional downstream tasks, such as dialogue generation. In addition, we wish to experiment further with different heuristics for defining domains. Within the scope of neural machine translation, we also intend to explore sentence filtering techniques to better leverage web-crawled Irish text.

Furthermore, we will continue experimenting with domain adaptation strategies, including combining SemiLoRA and SemiAdapt with a fully fine-tuned base model.
We hope that the introduction of SemiLoRA and SemiAdapt will empower researchers working on LRLs such as Irish to develop more accessible and efficient language technologies. These tools can help address technological language inequality, which is being intensified by computationally expensive LLM-based systems that predominantly benefit majority languages with disproportionate online representation.

\section{Limitations}
This work focused on higher-quality, non–web-crawled data sources; future work will evaluate the methods on additional datasets. Although a substantial portion of web-crawled text for Irish is not natural or linguistically correct, excluding it limits the available training data. We therefore plan to investigate quality assessment and filtering methods to better leverage web-crawled sentence pairs. Finally, we did not explore the Irish to English translation direction, as decoding into English is generally stronger in large multilingual models, but future work will address this.

\section{Ethics Statement}
This work contributes to the Irish language by open-sourcing new resources and demonstrating new parameter-efficient approaches (SemiLoRA and SemiAdapt) designed to empower researchers working on other LRLs. Our goal is to promote more equitable language technology development and to help address the technological language inequality reinforced by LLM-based systems trained on biased internet data.

% \section{Acknowledgements}
% This publication has emanated from research conducted with the financial support of Taighde Éireann - Research
% Ireland under Grant number 18/CRT/6223.

\nocite{*}
\nocitelanguageresource{*}   % cite everything in language resource .bib

\section{Bibliographical References}\label{sec:reference}

\bibliographystyle{lrec2026-natbib}
\bibliography{lrec2026-example}

\section{Language Resource References}
\label{lr:ref}
\bibliographystylelanguageresource{lrec2026-natbib}
\bibliographylanguageresource{languageresource}

\end{document}